\documentclass[11pt]{article}

\usepackage[preprint]{acl}
\usepackage{times}
\usepackage{latexsym}
\usepackage[T1]{fontenc}
\usepackage[utf8]{inputenc}
\usepackage{microtype}
\usepackage{inconsolata}
\usepackage{graphicx}
\usepackage{booktabs}
\usepackage{float}
\usepackage{needspace}
\usepackage{tabularx}
\usepackage{array}
\usepackage{xcolor}
\usepackage{tikz}
\usepackage{listings}
\usepackage{url}
\usetikzlibrary{arrows.meta,calc,positioning}

\lstdefinestyle{paperprompt}{
  basicstyle=\ttfamily\fontsize{8}{9.5}\selectfont,
  columns=fullflexible,
  breaklines=true,
  keepspaces=true,
  frame=single,
  framerule=0.4pt,
  framesep=3pt,
  aboveskip=4pt,
  belowskip=4pt
}

\title{Recognizing Study Lineage Is Not Enough: Over-Crediting Same-Trial Evidence in LLM-Based Medical Verification}

\author{
Mukand Krishna$^{1}$ \qquad
Anoushka Harit$^{2}$ \qquad
Chenghao Xiao$^{3}$ \qquad
Zhongtian Sun$^{1}$\thanks{Corresponding author: \texttt{z.sun-256@kent.ac.uk}} \\[2pt]
\normalfont\normalsize
$^{1}$School of Computing, University of Kent, UK \\
$^{2}$Cancer Research UK Cambridge Institute, University of Cambridge, UK \\
$^{3}$Shanghai University of Finance and Economics, China \\[2pt]
$^{1}$\texttt{\{mk116,z.sun-256\}@kent.ac.uk} \\
$^{2}$\texttt{anoushka.harit@cruk.cam.ac.uk} \qquad
$^{3}$\texttt{xiaochenghao@sufe.edu.cn}
}

\begin{document}
\raggedbottom
\maketitle

\begin{abstract} 

Publication count is not study count. A model can identify shared trial lineage correctly in a direct task yet over-credit a second publication from the same trial when aggregating evidence. We evaluate this behavior in medical LLM verification using 14 source-audited clinical families and five model conditions. The families span provenance, clinical applicability, result direction, and composition; the provenance families provide the direct lineage tests. Recognition and aggregation are separate stateless calls, and repeated calls assess within-input stability. Across all three original provenance families, GPT-5.5 identifies the independent update in 10/10 recognition calls per family but over-credits the same-trial report in 5/5 separate aggregation calls per family. Selected Qwen3-4B and Ministral-3-8B conditions show the same cross-task contrast. A study-grouping instruction reduces GPT-5.5 over-credit from 4/5 to 0/5 in each later pure-provenance family. Other model conditions show mixed behavior, and rejected outputs limit several comparisons. The results identify a behavioral gap between elicited lineage recognition and evidence aggregation. The code and data artifacts are available at \url{https://anonymous.4open.science/r/study-lineage-review-artifact/}.

\end{abstract}

\section{Introduction}

A single clinical trial can generate several publications: a primary report may be followed by subgroup, safety, or longer-term analyses. These papers may support the same claim, but they come from one study. Clinical datasets can retain source links between trial results and supporting publications; CliniFact, for example, derives claims from trial results and links them to supporting scientific publications \citep{zhang-etal-2025-dataset}. A verifier that treats documents as independent studies can overstate the independence of the evidence.

\begin{figure}[H]
\centering
\includegraphics[width=\columnwidth]{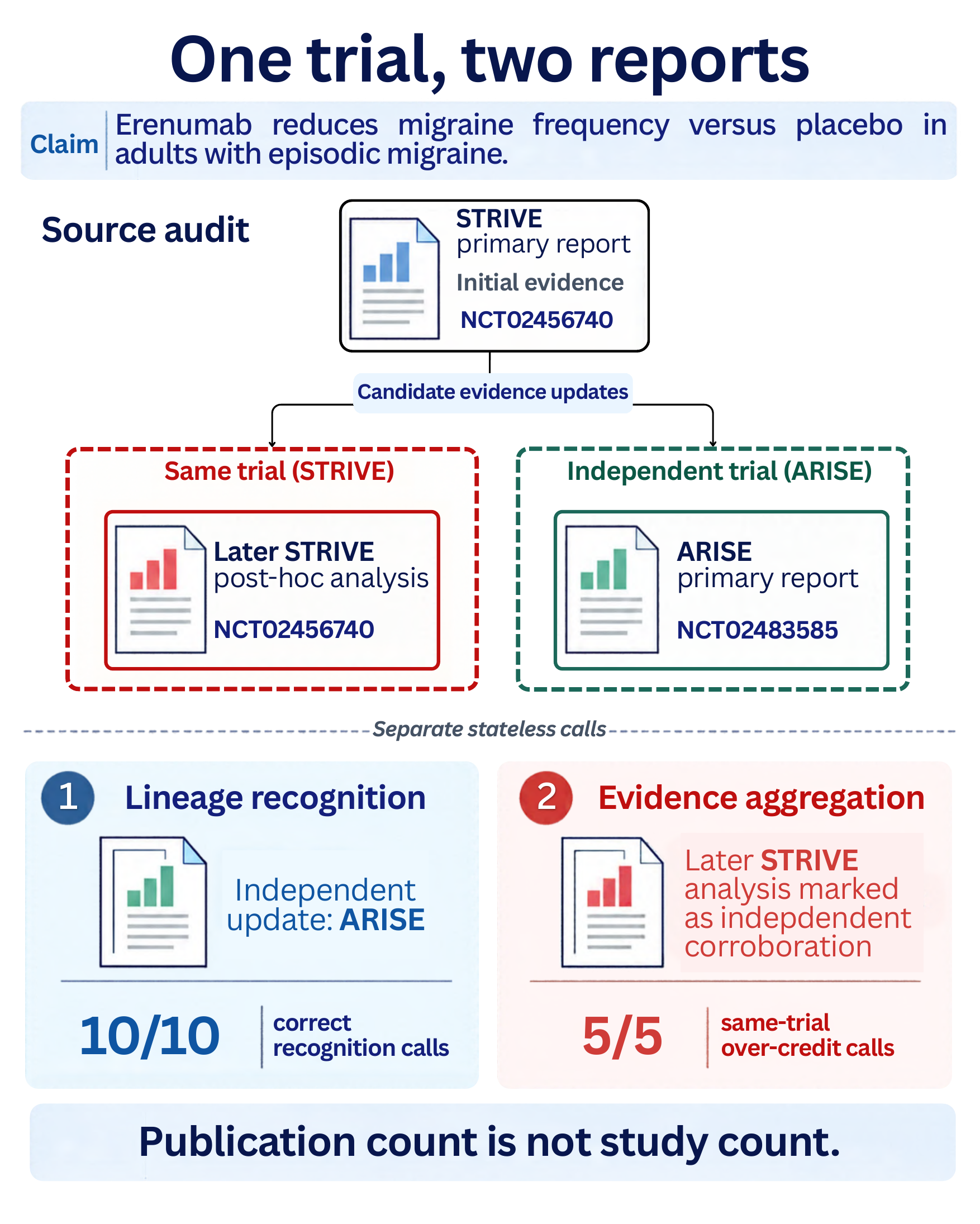}
\caption{A source-audited example of the recognition--aggregation gap. The STRIVE primary report and later analysis derive from trial NCT02456740, whereas the ARISE primary report comes from an independent trial (NCT02483585) \citep{goadsby-etal-2017-erenumab,diener-etal-2021-strive,dodick-etal-2018-arise}. In separate stateless GPT-5.5 calls, ARISE is identified as the independent update in 10/10 recognition calls (five per candidate order), but the later STRIVE analysis is marked as independent corroboration in 5/5 aggregation calls. Repeated calls assess within-input stability; the clinical family is the experimental unit.} \label{fig:strive-arise-gap}
\end{figure}

Existing evaluations clarify several parts of this problem. FACTPICO tests whether plain-language summaries preserve trial populations, interventions, comparators, outcomes, and findings \citep{joseph-etal-2024-factpico}. MediEval separates medical truth from support in a patient context \citep{qu-farber-2026-medieval}, while FORCEBENCH shows that a relevant citation may still be too weak or narrow for the claim attached to it \citep{qian-etal-2026-relevant}. Multi-document studies find that paraphrased repetition can outweigh distinct support and reverse source preferences under conflict \citep{naphade-2026-rational,schuster-etal-2026-whose}. ECON separates conflict detection from resolution \citep{jiayang-etal-2024-econ}. When Facts Change finds that a verbalized temporal distinction may fail to affect the final prediction \citep{wallat-etal-2026-facts}. These findings motivate a source-level test of whether an LLM verifier treats distinct publications from one clinical study as dependent evidence when combining their findings.

Our evaluation covers 14 source-audited clinical families across five model conditions. The provenance block directly tests whether an added publication derives from the same study or from an independent study. The remaining families test clinical applicability, result direction, and combinations of these relations. Within each family, we keep the claim fixed and vary the relation of the added evidence. Relation recognition and evidence aggregation are evaluated in separate stateless calls. Repeated calls assess within-input stability; clinical families are the experimental units.

The experiments address three connected questions. First, can a verifier identify shared study lineage? Second, when separate recognition calls are correct, do aggregation calls respect that dependence? Third, can explicit study-level grouping reduce same-trial over-credit while preserving support from a genuinely independent trial? The second question defines the paper's central behavioral contrast.

Across all three original provenance families, GPT-5.5 \citep{openai-2026-gpt55} answers the lineage comparison correctly but still over-credits the same-trial publication in the separate aggregation condition. Selected Qwen3-4B \citep{qwen-team-2025-qwen3} and Ministral-3-8B \citep{mistral-ai-2025-ministral3} conditions show the same family-level contrast. Explicit study grouping removes the error for GPT-5.5 in both later pure-provenance families while preserving support for the independent trial. Grouping effects vary across models, and rejected outputs leave some comparisons unresolved.

The paper makes three contributions:
\begin{itemize}
    \item We show a family-level recognition--aggregation gap: a model can correctly identify shared study lineage in a direct task yet still over-credit a same-trial publication in a separate evidence-aggregation task.

    \item We construct a source-audited evaluation of 14 clinical families that separates relation recognition from aggregation and treats each clinical family as the experimental unit and repeated calls as stability checks.

    \item We test whether an explicit study-grouping intervention reduces same-trial over-credit while preserving support for a genuinely independent trial.
\end{itemize}

\section{Related work}

\subsection{Medical evidence verification and contextual grounding}

Fine-grained factuality evaluators operate below the whole-answer level. FActScore decomposes long-form generations into atomic facts and measures the proportion supported by a knowledge source \citep{min-etal-2023-factscore}. ClaimVer identifies individual claims, verifies them against a knowledge graph, and returns evidence and an explanation for each prediction \citep{dammu-etal-2024-claimver}. This granularity matters in medicine because an answer may contain both supported and unsupported clinical statements.

Medical benchmarks make the claim--evidence relation more explicit. FACTPICO evaluates whether plain-language summaries of randomized controlled trials preserve the population, intervention, comparator, outcome, and reported findings \citep{joseph-etal-2024-factpico}. Clinical-trial NLI tests entailment and contradiction over trial reports, including controlled perturbations \citep{jullien-etal-2024-semeval}. MediEval crosses medical truth with support in a specific patient record \citep{qu-farber-2026-medieval}, and CliniFact derives clinical claims from trial results and links them to supporting publication evidence \citep{zhang-etal-2025-dataset}. FORCEBENCH tests a related boundary in cited RAG: a passage can be relevant yet fail to warrant a claim's relation, certainty, scope, temporal status, or numeric precision \citep{qian-etal-2026-relevant}. These clinical claim-verification evaluations do not test whether aggregation respects trial-level dependence across publications.

\subsection{Source dependence and recognition--use gaps}

Clinical evidence synthesis treats the study, rather than its reports, as the unit of interest. The Cochrane Handbook calls for multiple reports of one study to be linked while retaining useful information from each \citep{li-etal-2024-cochrane}. Aggregator uses PubMed metadata to cluster articles likely to report the same registered trial \citep{smalheiser-holt-2020-aggregator}. That work addresses the linkage step; our evaluation asks whether an LLM verifier respects that linkage when combining evidence for a claim.

Multi-document studies show that repetition can change model behavior without adding an independent source. GroupQA finds that paraphrased repetition can be more persuasive than distinct supporting documents \citep{naphade-2026-rational}. Whose Facts Win? shows that repeating information from a less-preferred source can reverse source preferences under conflict \citep{schuster-etal-2026-whose}. Bara formalizes a related distinction between report multiplicity and evidential ancestry in controlled multi-agent inference \citep{bara-2026-epistemic-sybil}. Our setting differs from literal or paraphrased duplication and from synthetic agent reports: the documents are distinct clinical publications, but some derive from the same clinical trial.

Other studies examine whether models act consistently on relations they can identify. ECON evaluates conflict detection separately from conflict resolution \citep{jiayang-etal-2024-econ}. Belief-R tests whether models revise a conclusion when a new premise requires an update and retain it otherwise \citep{wilie-etal-2024-belief}. When Facts Change reports that larger models can verbalize the relevant temporal distinction without consistently carrying it into their final predictions \citep{wallat-etal-2026-facts}. Sufficient Context similarly separates whether retrieved context contains enough information from whether a model uses it correctly \citep{joren-etal-2025-sufficient}. These studies motivate our separation of lineage recognition from evidence aggregation, although our two tasks are evaluated in independent model calls.

Adjacent medical-AI work examines when text-side bias can help medical visual question answering and how adversarial image--text conflicts can disrupt multimodal medical RAG \citep{sun2022unimodal,zuo2025make}. Recent work on monosemantic attribution instead seeks more stable representation-level explanations for clinical language models \citep{mamalakis2026monosemantic}. Our study remains behavioral: it tests whether a verifier's answers across separate recognition and aggregation tasks respect the same source relation.

\section{Task formulation}
\label{sec:task}

\subsection{Evidence-update setting}

For a fixed medical claim $c$, an initial evidence state $E_0$ receives one source-linked update $u$, producing $E_1=E_0\cup\{u\}$. Each update isolates one target relation. \emph{Provenance} asks whether the added report is an exact repeat, another publication from the same underlying study, or evidence from an independent study. \emph{Applicability} asks whether it matches the claim's population, intervention, comparator, outcome, and time frame. \emph{Direction} asks whether the observed claim-relevant result agrees with or challenges the claim. Study independence alone does not establish corroboration: an independent result must also apply to and support the fixed claim. A same-study report can add relevant detail without adding a second independent study. Positive evidence outside the claim's scope should leave the scoped judgment unchanged, while applicable opposing evidence should register conflict without erasing earlier support.

The provenance decision is about the ancestry of the evidence, rather than the number of passages or papers in the input. Let $S(E)$ be the set of underlying studies represented by evidence state $E$, and let $s(u)$ be the study that produced update $u$. A later analysis with $s(u)\in S(E_0)$ leaves the number of studies unchanged even when it reports a new outcome or has a new publication identifier. An update with $s(u)\notin S(E_0)$ adds a study, but it counts as corroboration only when its result is applicable and agrees with the claim. This distinction matters for the combined judgment: two supportive passages can warrant a \texttt{supported} label without warranting an \texttt{independent\_corroboration} flag.

The held-out H3 family puts these relations around one fixed statin claim. Its updates include a later WOSCOPS analysis from the same trial lineage, positive evidence from 4S that falls outside the claim's clinical scope, and a challenging result from ALLHAT-LLT \citep{shepherd-etal-1995-woscops,vallejo-vaz-etal-2017-woscops,scandinavian-1994-4s,allhat-2002-llt}. We use H3 as a compositional test that combines relations within one family. Appendix Table~\ref{tab:family-inventory} records the source identities used in this family.

\subsection{Lineage recognition and evidence aggregation}

We evaluated recognition and aggregation in separate stateless calls. The recognition task presents the same initial evidence with two candidate updates and asks which one satisfies a single target relation; both A/B and B/A orders are tested. The aggregation task receives the current evidence state and returns a combined judgment. Its input excludes both the expected relation and the recognition response.

We ran five fresh invocations for each exact input. Relation recognition and relation-specific aggregation behavior are the primary outcomes; numerical support and confidence scores are diagnostic. Outputs that fail the response contract are reported as rejected and receive no semantic score. Appendix~\ref{app:prompt-contract} provides the full prompts, response schemas, validation rules, and execution details.

\subsection{Evaluation outcomes and validity}

The primary contrast asks whether correct lineage answers coexist with appropriate handling in the separate aggregation task. We define same-trial over-credit as a schema-valid aggregation output that marks the added report as independent corroboration even though the source audit links it to a study already represented in the initial evidence. We audit independent-trial preservation separately: the output must both mark independent corroboration and use the added independent evidence. Repeated calls estimate within-input stability, while clinical families remain the experimental cases. Rejected outputs remain in planned-call accounting but receive no semantic score.

The combination flag is the primary measure of over-credit, rather than a change in the 0--100 support score. A later report can legitimately add detail and affect the model's assessment without becoming a second independent study. Conversely, the overall claim can remain \texttt{supported} while the model misstates how many independent studies support it. For the independent-trial check, requiring both the flag and use of the added item distinguishes genuine incorporation from a flag attached to an ignored passage. We apply these categorical rules after validating the response structure and evidence quotations.

\section{Benchmark construction}
\label{sec:benchmark}

\subsection{Source-audited clinical families}

Each clinical family contains one fixed claim, an initial evidence set, matched updates that isolate a declared relation, and a null control. We established the relation labels before model inference. For provenance, we cross-checked publication records, preserved source metadata, and trial identifiers where available to determine whether reports shared an underlying study. We audited applicability against the claim's clinical scope and assigned direction according to the observed claim-relevant result, with statistical significance considered as supplementary context. The audit also retained the exact passages and source locators used in each input.

We accepted a family only when the initial evidence already supplied the information needed to assess the fixed claim, the candidate updates isolated one clinically material relation, and the source relation could be checked independently. An update that supplied a missing part of the claim or changed several clinical dimensions at once would confound the intended contrast. For provenance, publication identifiers traced reports, while trial records and study descriptions established whether those reports came from the same study. After acceptance, we froze the claim, passages, relation labels, and expected categorical handling before the corresponding model calls. Appendix~\ref{app:family-inventory} gives the complete family inventory, source identities, audit chronology, and analysis status.

The controls are designed to separate errors that can look alike in a final verdict. In CF-A3, a passage about olaparib maintenance after relapse reports a positive progression-free-survival result, but the fixed claim concerns first-line maintenance after platinum response. A verifier can read the result correctly and still use it for the wrong population. CF-D3 makes a different demand: the initial evidence reports fewer falls with vitamin D plus calcium, whereas the added annual high-dose vitamin-D trial reports more falls. The latter passage is relevant to the claim and should register a challenge rather than be ignored. These examples distinguish scope and direction errors from the provenance error tested in CF-P1--P3; the selected passage identities are traceable in Appendix Tables~\ref{tab:family-inventory} and~\ref{tab:initial-source-anchors}.

CF-P2 illustrates how the provenance contrast is constructed. The initial passages describe the QUEST dupilumab trial (\texttt{NCT02414854}). One candidate is a later analysis of QUEST that examines outcomes by exacerbation history; its different title and PMID still link it to the same trial. The other candidate reports a separate dupilumab trial (\texttt{NCT01854047}) with claim-relevant exacerbation and lung-function findings. Both candidates concern add-on dupilumab in uncontrolled asthma, so choosing between them is a lineage decision rather than a choice between a positive and a negative result. The verifier then receives each update separately, making it possible to check whether it reserves independent corroboration for the second trial.

\subsection{Experimental unit and freezing}

The clinical family is the experimental unit because its matched conditions share one claim and one audited contrast. Repeated calls reuse the same input and characterize response stability within a family. Independent-study status required corroborating source records, metadata, and trial identifiers. The original and follow-up suites were frozen separately, so we report them as distinct stages, preserving their separate freezing histories.

\subsection{Evaluation suites and analysis status}

We ran the evaluation in stages. The original, prospectively frozen suite contains three provenance families (CF-P1--P3), three applicability families (CF-A1--A3), three direction families (CF-D1--D3), and the held-out compositional family H3. A separately frozen follow-up package adds two pure-provenance families (P5R and P6R) and two compositional families (H5 and H6). Table~\ref{tab:benchmark-overview} lists their clinical targets and contrasts.

\begin{table*}[t]
\centering
\small
\setlength{\tabcolsep}{4pt}
\renewcommand{\arraystretch}{1.08}
\begin{tabularx}{\textwidth}{@{}l >{\raggedright\arraybackslash}X >{\raggedright\arraybackslash}X c@{}}
\toprule
\textbf{Family} & \textbf{Clinical target} & \textbf{Tested contrast} & \textbf{Inputs} \\
\midrule
CF-P1 & Erenumab; episodic migraine & STRIVE post-hoc analysis vs ARISE & 6 \\
CF-P2 & Dupilumab; uncontrolled asthma & QUEST analysis vs distinct trial & 6 \\
CF-P3 & SGLT2 inhibition; kidney disease & EMPA-KIDNEY analysis vs distinct trial & 6 \\
CF-A1 & Adjuvant pembrolizumab; melanoma & applicable vs out-of-scope evidence & 6 \\
CF-A2 & Factor Xa inhibition; atrial fibrillation & applicable vs out-of-scope evidence & 6 \\
CF-A3 & Olaparib; ovarian-cancer maintenance & first-line vs relapsed disease & 6 \\
CF-D1 & ICU glucose control & agreeing vs mortality-challenging result & 6 \\
CF-D2 & Intensive glycemic control & agreeing vs challenging outcome & 6 \\
CF-D3 & Vitamin D; falls in older adults & fall reduction vs high-dose challenge & 6 \\
H3 & Pravastatin; coronary-event prevention & provenance, scope, and direction & 12 \\
\midrule
P5R & Early antiretroviral therapy & START follow-up vs TEMPRANO; grouping & 10 \\
P6R & Checkpoint inhibitor plus chemotherapy; lung cancer & same-trial vs independent update; grouping & 10 \\
H5 & Antiretroviral treatment evidence & compositional follow-up & 14 \\
H6 & Checkpoint inhibitor evidence & compositional follow-up & 10 \\
\midrule
\textbf{Total} & \textbf{14 clinical families} & & \textbf{110} \\
\bottomrule
\end{tabularx}
\caption{Clinical targets and analytical roles of the selected 14 families. Inputs are distinct task inputs per model, each invoked five times. The first ten families belong to the original suite (66 inputs); the final four were frozen later (44 inputs). Source identifiers and publication anchors appear in Appendix Tables~\ref{tab:family-inventory} and~\ref{tab:initial-source-anchors}.}
\label{tab:benchmark-overview}
\end{table*}

Oracle and cue-ablation tests were introduced after the primary provenance failure and are treated as post-hoc diagnostics. The prompt-order audit is secondary, and the lineage-by-overlap family is exploratory and policy-instructed. These analyses are not pooled with the primary 14-family comparison. Appendix~\ref{app:study-chronology} records the experimental chronology.

\section{Experiments and results}
\label{sec:results}

\subsection{Experimental setup}

We evaluated five model conditions. \nolinkurl{gpt-5.5} used medium reasoning effort. The open-weight models were \nolinkurl{Qwen3-4B-Instruct-2507}, \nolinkurl{Gemma-3-4B-it} \citep{gemma-team-2025-gemma3}, \nolinkurl{Llama-3.1-8B-Instruct} \citep{dubey-etal-2024-llama3}, and \nolinkurl{Ministral-3-8B-Instruct-2512}. The controls cover exact repetition, same-trial versus independent-study updates, positive but out-of-scope evidence, applicable opposing evidence, and counterbalanced recognition order. Primitive confirmatory inputs omit the study-group label used in the grouped condition; H3 remains a cue-rich compositional reference.

The open-weight runs used greedy decoding (\texttt{do\_sample=false}, \texttt{max\_new\_tokens=2048}). In the later grouping condition, the verifier assessed all displayed reports with study identifiers and an instruction to combine evidence by underlying trial identity.

Both tasks returned structured responses, which we checked before reading the clinical judgment. Validation covered the response schema, the requested item identity, coverage and order of displayed evidence, and exact-substring quotations from the passages. We retained raw outputs and validation errors; canonical validity permits removal of one outer Markdown JSON fence while leaving the response content unchanged. Each candidate order in recognition was invoked five times. The grouping condition retains all displayed documents and adds a study-level aggregation instruction. Appendix~\ref{app:prompt-contract} records the complete templates and model-specific formatting details.

Validity differs substantially across models. Of 550 planned calls per model in the selected 14-family comparison, GPT-5.5 has 550 canonical-valid outputs, Qwen 480, Gemma 400, Llama 395, and Ministral 410. These totals describe response validity across recognition, controls, and aggregation. A family with five rejected aggregation calls contributes no estimate of that outcome, even when its recognition answers are correct. Rejected cells appear alongside the family-level semantic results (Appendix Table~\ref{tab:family-validity}).

\subsection{Models can recognize study lineage directly}

Across the three original provenance families, GPT-5.5 and Qwen identify the independent update in all ten recognition calls in each family. Gemma and Llama are correct in five of ten calls per family. Ministral reaches 10/10 in CF-P1; in CF-P2 and CF-P3, five calls are correct and five are rejected. Appendix Table~\ref{tab:original-provenance-detail} gives the full family-level counts.

The two recognition orders matter to the interpretation. Each original provenance family supplies five calls with one candidate as A and five with the candidates reversed; full recognition requires the correct study relation in both presentations. A 5/10 result therefore falls short of the order-balanced criterion even if every valid answer in one presentation is correct. Recognition asks for the independent update, whereas verification assesses the combined support for the claim.

\subsection{Correct lineage recognition does not guarantee lineage-respecting aggregation}

Correct lineage answers and lineage-respecting aggregation diverge. 
GPT-5.5 fully recognizes all three original provenance families, but the separate aggregation task over-credits the same-trial update in 5/5 calls in each family. Qwen shows the same contrast in CF-P1 and CF-P3; its CF-P2 aggregation outputs are rejected. Ministral shows the contrast in CF-P1. Gemma and Llama do not achieve full two-order recognition in these families. Table~\ref{tab:provenance-results} reports the family-level contrast; recognition responses are not inputs to aggregation calls.

The exact-repeat and null controls sharpen this finding. Across the nine original primitive families, GPT-5.5 treated these controls as duplicate or unchanged in 45/45 calls. In CF-P1--P3, it marked the later same-trial publication as independent corroboration in 15/15 calls. The model thus handled literal repetition differently from a second publication linked to an already represented trial. The 15 calls are repeated measurements across three clinical families.

\begin{table}[t]
\centering
\small
\setlength{\tabcolsep}{2pt}
\renewcommand{\arraystretch}{1.08}
\begin{tabularx}{\columnwidth}{@{}>{\raggedright\arraybackslash}X >{\centering\arraybackslash}p{0.16\columnwidth} >{\centering\arraybackslash}p{0.29\columnwidth} >{\centering\arraybackslash}p{0.14\columnwidth}@{}}
\toprule
\textbf{Model} &
\shortstack{\textbf{Recognition}} &
\shortstack{\textbf{Aggregation}} &
\shortstack{\textbf{Rejected}\\\textbf{R/A}} \\
\midrule
GPT-5.5        & 3/3 & 0/3 & 0/0 \\
Qwen3-4B       & 3/3 & 0/2 & 0/5 \\
Gemma3-4B      & 0/3 & \textit{N/E} & 0/10 \\
Llama-3.1-8B   & 0/3 & \textit{N/E} & 0/10 \\
Ministral-3-8B & 1/3 & 0/1 & 10/5 \\
\bottomrule
\end{tabularx}
\caption{Family-level recognition--aggregation contrast in CF-P1--P3.
Full recognition requires 10/10 correct calls across both
counterbalanced orders. The aggregation column counts
lineage-respecting families among fully recognized families with valid
same-trial aggregation outputs.
N/E means not evaluable because no family met this prerequisite. A
family is lineage-respecting when the added same-trial report is not
labeled as independent corroboration. R/A denotes rejected
recognition/aggregation calls.}
\label{tab:provenance-results}
\end{table}

\begin{figure*}[t]
\centering
\includegraphics[width=\textwidth]{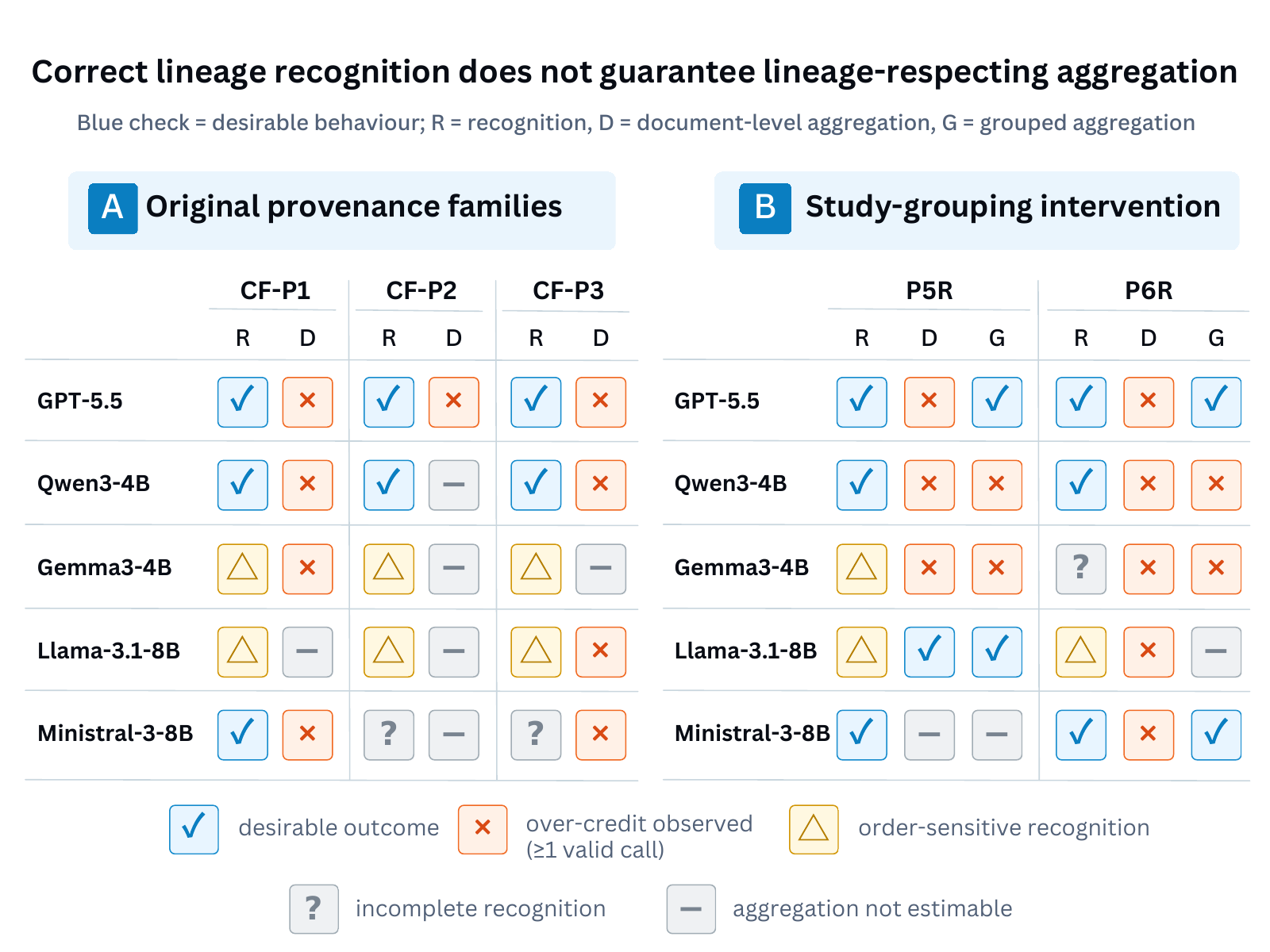}
\caption{Model-by-family results for lineage recognition and same-trial evidence aggregation. Panel A reports the original provenance families; Panel B reports the later pure-provenance families used to test study grouping. R denotes full recognition across both candidate orders (10/10 correct calls); D and G indicate whether document-level and study-grouped aggregation, respectively, avoided same-trial over-credit in every valid call. A blue check marks that the criterion was met; an orange cross indicates over-credit in at least one valid call. A triangle marks order-sensitive recognition, a question mark marks recognition limited by rejected outputs, and a dash indicates that aggregation is not estimable. Preservation of genuinely independent evidence is evaluated separately in Appendix Table~\ref{tab:later-detail}.}
\label{fig:grouping-results}
\end{figure*}

\subsection{Study grouping repairs some model--family conditions}

P5R is the clearest intervention case. Its initial START report and later START follow-up derive from the same trial, whereas TEMPRANO is an independent trial \citep{lundgren-etal-2015-start,lundgren-etal-2023-start-followup,danel-etal-2015-temprano}. GPT-5.5 recognizes this distinction in 10/10 calls but marks the later START report as independent corroboration in 4/5 document-level calls. Under study grouping, over-credit falls to 0/5. Qwen and Gemma remain at 5/5 over-credit after grouping. Llama has no document-level over-credit in P5R, while Ministral has no valid same-trial document or grouped output for this family.

P6R provides a second pure-provenance comparison. GPT-5.5 again moves from 4/5 document-level over-credit to 0/5 under grouping, and Ministral moves from 5/5 to 0/5. Qwen and Gemma remain at 5/5. Rejected grouped outputs leave the Llama comparison unresolved.

The independent-trial update provides a separate check on the grouping intervention. In P6R, all five models mark and use it in 5/5 calls before and after grouping. GPT-5.5, Qwen, Gemma, and Llama show the same 5/5 pattern in P5R; Ministral's grouped outputs for that update are rejected. GPT-5.5's reduction in same-trial over-credit therefore coincides with continued use of independent evidence in both families (Appendix Table~\ref{tab:later-detail}).

H5 and H6 give a different view of the grouping instruction. For GPT-5.5, the global corroboration flag on the same-trial-update condition changes from 5/5 to 4/5 in H5 and from 5/5 to 1/5 in H6. Both initial evidence states already contain an independent study, so a global flag cannot identify which study received the credit. We report those changes as compositional observations and reserve the same-trial over-credit comparison for P5R and P6R (Appendix Table~\ref{tab:compositional-detail}).

\section{Further analysis}
\label{sec:further-analysis}

\subsection{Specificity and compositional evidence}

Applicability and direction provide specificity controls for the central lineage finding. GPT-5.5 recognizes the target update in all six CF-A/CF-D families (10/10 per family). In separate aggregation calls, it ignores the positive but out-of-scope update in each applicability family and registers the challenging update as conflicting or mixed evidence in each direction family (5/5 per family).

H3 asks whether the same pattern survives when the relations are no longer isolated. The family centers on one statin claim and includes a later WOSCOPS analysis from the same lineage, positive but out-of-scope evidence from 4S, and a challenging result from ALLHAT-LLT. GPT-5.5 and Qwen are correct in all 30 recognition calls across provenance, applicability, and direction. Yet their separate aggregation calls still mark the added same-trial WOSCOPS analysis as independent corroboration in 4/5 and 5/5 valid calls, respectively. Those 30 calls comprise ten comparisons for each of the three relations. Full control-axis results are reported in Appendix~\ref{app:control-axis-results}.

\subsection{Explicit provenance information repairs one model condition}

CF-P3 tests whether externally supplied provenance information changes aggregation behavior. GPT-5.5 over-credits the same-trial EMPA-KIDNEY analysis in 5/5 neutral-verifier calls. When the verifier is given either the item-specific relation or a general study-dependence policy, the same-trial independent-corroboration flag disappears in all five valid calls, while the independent-trial control remains at 5/5. Qwen retains the same-trial independent-corroboration flag in all five calls under both oracle conditions. Explicit provenance information therefore repairs this GPT-5.5 condition, while Qwen shows the same
over-credit pattern under the two oracle conditions.

This diagnostic evaluates externally supplied relation and policy information. Testing whether a model's own recognition response improves later aggregation requires a separate two-stage design. H3-M1 tests a different cue. After removing the study-group cue, GPT-5.5 still recognizes provenance in 10/10 calls and over-credits the same-trial report in 5/5 separate aggregation calls.  Appendix~\ref{app:diagnostic-results} gives the full model-specific denominators.

\subsection{Prompt-order and lineage-by-overlap controls}

To test sensitivity to the recorded verifier wording, we repeated the P5R/P6R same-trial and independent-update conditions with a semantically equivalent prompt whose instructions were reordered. Under canonical parsing, same-trial over-credit still appeared in 16/20 GPT-5.5 records and in all 20 Qwen3-4B and Gemma3-4B records. The independent-update audit was less uniform (14/20, 15/20, and 20/20, respectively), and Gemma's raw replies contained a removable outer Markdown fence. This secondary audit therefore preserves the broad pattern under the two stored prompt variants, but it does not estimate general prompt robustness.

The exploratory ProMISe/ARISE control addresses a different question. It compares source-faithful same-trial and independent-trial passages at higher and lower overlap with the initial anchor, while explicitly stating the study-level policy in the verifier prompt. GPT-5.5 produced the expected handling in all 20 calls. This shows policy application across the four fixed cells, rather than unprompted lineage discovery. Because the high- and low-overlap passages are selected strata within one family, not randomized treatments, the control does not identify a causal effect of textual overlap. Appendix~\ref{app:secondary-analyses} provides the frozen passages, overlap measurements, and full outcome details.

\section{Conclusion}

The results show a cross-task behavioral gap: study lineage can be recognized in a direct comparison yet fail to guide a separate evidence-aggregation judgment. Across the three original provenance families, GPT-5.5 consistently identified the independent trial but treated a later report from the same trial as independent corroboration. Selected Qwen3-4B and Ministral-3-8B conditions showed the same cross-task pattern. In the later pure-provenance families, study grouping eliminated GPT-5.5 same-trial over-credit while retaining support for independent evidence; effects varied across the other model conditions.

These results show why evidence aggregation should use underlying study identity as the unit of corroboration. The evidence comes from 14 source-audited clinical families, with repeated calls characterizing within-input stability. The staged design distinguishes the original suite, the later grouping follow-up, and post-hoc and exploratory diagnostics. Its scope is the audited families and the displayed source metadata; broader prevalence and cue-free grouping require further evaluation. The next question is whether models can infer the same grouping from ordinary publication metadata.

\clearpage

\section*{Limitations}

The 14 source-audited families are selected clinical contrasts, not a representative sample of medical claims or publications. Repeated calls measure within-input stability rather than independent clinical cases, and separate stateless recognition and aggregation calls do not reveal the model's internal representation. The grouping condition exposes study identifiers and changes the prompt, while rejected outputs limit some cross-model comparisons; these boundaries constrain generalisation and causal interpretation.

\bibliography{custom}


\appendix
\raggedbottom

\section{Clinical families, source identity, and study chronology}
\label{app:family-inventory}

\subsection{Ground truth, freezing, and experimental unit}
\label{app:study-chronology}

We established study lineage from the source record before model inference. Where available, trial registration identifiers provided the clearest link between publications; we also retained publication titles, PMIDs or other identifiers, source locators, and the study descriptions in the frozen records. We assigned a same-trial label only when the selected reports could be linked to the same underlying study; an independent label required a distinct trial or cohort. We audited applicability and direction separately against the fixed claim's population, intervention, comparator, outcome, time frame, and observed result.

For each accepted family, we froze the claim, initial passages, candidate updates, source metadata, relation labels, expected categorical handling, and source-audit records before the corresponding calls. The original ten-family suite and the later four-family package were frozen at different stages. The later P6R source correction is documented in Appendix~\ref{app:prompt-contract}; superseded and selected inputs were not pooled. The clinical family is the experimental unit because its matched conditions share one claim and one audited contrast. Repeated calls of an exact input measure response stability only.

\begin{table*}[t]
\centering
\small
\begin{tabularx}{\textwidth}{@{}l l X X@{}}
\toprule
Design component & Families/conditions & Analysis status & Role in the paper \\
\midrule
Original suite &
\begin{tabular}[t]{@{}l@{}}
CF-P1--P3, CF-A1--A3,\\
CF-D1--D3, H3
\end{tabular} &
prospectively frozen primary evaluation &
recognition and aggregation across provenance, applicability, direction, and composition \\
Later package & P5R, P6R, H5, H6 & separately frozen follow-up & study grouping and preservation of genuinely independent evidence \\
Prompt-order audit & P5R, P6R & secondary robustness & tests one semantically equivalent verifier-prompt reordering \\
Oracles and cue ablation & CF-P3 oracles, H3-M1 & post-hoc diagnostic & tests explicit relation/policy information and removal of the study-group cue \\
Lineage $\times$ overlap & ProMISe/ARISE control & exploratory, policy-instructed & separates application of a stated lineage rule from ordinary passage overlap \\
\bottomrule
\end{tabularx}
\caption{Experimental chronology and analysis status. Later diagnostics were not part of the original primary denominator and are reported separately.}
\label{tab:study-chronology}
\end{table*}

\subsection{Selected family inventory}

Table~\ref{tab:family-inventory} lists the 14 families in the selected comparison. The clinical targets provide abbreviated labels for the tested claims. Study identifiers are transcribed from the selected frozen verifier inputs; a missing identifier is not evidence that two reports describe the same study.

\begin{table*}[t]
\centering
\footnotesize
\begin{tabularx}{\textwidth}{@{}l c X X@{}}
\toprule
Family & Inputs & Tested claim/contrast & Study identifiers recorded in selected verifier inputs \\
\midrule
CF-P1 & 6 & Erenumab and episodic-migraine frequency & STRIVE: \texttt{NCT02456740}; ARISE: \texttt{NCT02483585} \\
CF-P2 & 6 & Add-on dupilumab in uncontrolled asthma & \texttt{NCT02414854}; \texttt{NCT01854047} \\
CF-P3 & 6 & SGLT2 inhibition in chronic kidney disease & \texttt{NCT03594110}; \texttt{NCT03036150} \\
CF-A1 & 6 & Adjuvant pembrolizumab and melanoma recurrence & \texttt{NCT02362594}; S1404; \texttt{NCT01866319} \\
CF-A2 & 6 & Factor Xa inhibition for atrial-fibrillation stroke prevention & \texttt{NCT00412984}; \texttt{NCT00403767}; \texttt{NCT00643201} \\
CF-A3 & 6 & First-line olaparib-containing ovarian-cancer maintenance & \texttt{NCT01844986}; \texttt{NCT02477644}; \texttt{NCT00753545} \\
CF-D1 & 6 & ICU glucose control and mortality & Leuven surgical ICU trial; \texttt{NCT00115479}; \texttt{NCT00220987} \\
CF-D2 & 6 & Intensive glycemic control and death or cardiovascular risk & \texttt{ISRCTN75451837}; \texttt{NCT00000620}; \texttt{NCT00145925} \\
CF-D3 & 6 & Vitamin-D-based supplementation and falls & \texttt{ACTRN12605000658617} for one source; other source records lack a trial ID \\
H3 & 12 & Pravastatin and coronary events in primary prevention & WOSCOPS; 4S; ALLHAT-LLT; \texttt{NCT00336336} \\
\midrule
P5R & 10 & Immediate versus deferred ART and serious events/death & START: \texttt{NCT00867048}; TEMPRANO: \texttt{NCT00495651} \\
P6R & 10 & Checkpoint inhibitor plus chemotherapy in metastatic NSCLC & \texttt{NCT02578680}; \texttt{NCT02367781} \\
H5 & 14 & Compositional ART evidence updates & \texttt{NCT00867048}; \texttt{NCT00495651}; \texttt{NCT00461513} \\
H6 & 10 & Compositional checkpoint-inhibitor evidence updates & \texttt{NCT02578680}; \texttt{NCT02366143}; \texttt{NCT02657434} \\
\bottomrule
\end{tabularx}
\caption{Selected family inventory. P, A, and D isolate provenance, applicability, and direction; H families combine relations. ``Inputs'' counts distinct task inputs per model, each invoked five times. The first ten rows are the original suite (66 inputs; 330 calls per model); the final four are the later package (44 inputs; 220 calls per model). Listed identifiers cover the selected verifier inputs, not necessarily every publication or every comparison-task source.}
\label{tab:family-inventory}
\end{table*}

Table~\ref{tab:initial-source-anchors} provides one traceable publication anchor for each initial state. Source labels and PMIDs are copied from the frozen input metadata. Where an initial state contains several passages or studies, the table shows one anchor only; Table~\ref{tab:family-inventory} records the wider identity set.

\begin{table*}[t]
\centering
\footnotesize
\begin{tabularx}{\textwidth}{@{}l X l@{}}
\toprule
Family & One initial-source label in the frozen input & PMID \\
\midrule
CF-P1 & A Controlled Trial of Erenumab for Episodic Migraine & 29171821 \\
CF-P2 & Dupilumab Efficacy and Safety in Moderate-to-Severe Uncontrolled Asthma & 29782217 \\
CF-P3 & Empagliflozin in Patients with Chronic Kidney Disease & 36331190 \\
CF-A1 & Adjuvant pembrolizumab versus placebo in resected stage III melanoma & 33857412 \\
CF-A2 & Apixaban versus warfarin in patients with atrial fibrillation & 21870978 \\
CF-A3 & SOLO1/GOG 3004 maintenance olaparib in newly diagnosed advanced ovarian cancer & 34715071 \\
CF-D1 & Intensive insulin therapy in critically ill patients & 11794168 \\
CF-D2 & Post-trial monitoring of a randomised controlled trial of intensive glycemic control in type 2 diabetes extended from 10 years to 24 years (UKPDS 91) & 38772405 \\
CF-D3 & Effects of vitamin D and calcium supplementation on falls & 12568412 \\
H3 & Prevention of coronary heart disease with pravastatin in men with hypercholesterolemia & 7566020 \\
\midrule
P5R & Initiation of Antiretroviral Therapy in Early Asymptomatic HIV Infection & 26192873 \\
P6R & Pembrolizumab Plus Chemotherapy in Metastatic Non-Small-Cell Lung Cancer & 29658856 \\
H5 & Initiation of Antiretroviral Therapy in Early Asymptomatic HIV Infection & 26192873 \\
H6 & Pembrolizumab Plus Chemotherapy in Metastatic Non-Small-Cell Lung Cancer & 29658856 \\
\bottomrule
\end{tabularx}
\caption{One initial-source anchor per family. These metadata identify the archived test inputs; they do not substitute for checking every added publication, and the tested claim is not asserted here as a current clinical recommendation.}
\label{tab:initial-source-anchors}
\end{table*}

\section{Prompts, execution, and implementation details}
\label{app:prompt-contract}

Figures~\ref{fig:recognition-prompt}--\ref{fig:grouped-prompt} reproduce the core templates: comparative recognition (v2.2), document-level verification (v2), and study-grouped verification (v1). These short names correspond to the full filenames and hashes in the frozen run configurations. The literal \texttt{\{model\_input\_json\}} placeholder is replaced, once per call, by the frozen JSON input containing the claim, displayed source metadata, and evidence passages. Gold answers and expected score movements are not in that input. The figures show the base templates; they omit the varying clinical JSON payload and the model-specific JSON-schema formatting suffix described below. Recognition answers are never passed to verifier calls.

\subsection{Inputs, output validation, and execution}
\label{app:prompt-execution}

The recognition input contains \texttt{claim\_text}, \texttt{target\_relation}, \texttt{initial\_evidence}, \texttt{update\_a}, and \texttt{update\_b}; verifier inputs contain the fixed claim, \texttt{initial\_evidence}, and either an \texttt{added\_evidence} object or null. Each displayed evidence object carries an evidence ID, ordered bundle position, source title, passage, and available source identifiers. Some original-suite conditions omit the study-group label (\texttt{source\_group\_id}); the later grouped condition includes it and \texttt{trial\_id}. This condition changes the aggregation prompt while retaining the displayed documents and source metadata.

Recognition responses must contain the item and target relation, an A/B/no-difference answer, confidence, metadata basis, exact passage quotations, and a reason. Verifier responses must assess every displayed item in order and return an overall label, scores, used/ignored evidence IDs, combination flags, and a reason. The reported over-credit measure reads the \texttt{independent\_corroboration} flag on the same-trial-update condition. We validated the JSON schema, item identity, evidence coverage and order, and quoted substrings before semantic scoring; rejected outputs remain in the denominator accounting but receive no semantic score.

GPT-5.5 ran through Codex with medium reasoning effort; that interface did not specify temperature or random seed. The open-weight runs used each model's chat template and greedy decoding (\texttt{do\_sample=false}, \texttt{max\_new\_tokens=2048}). In the combined later-package Qwen/Gemma runner, a provider adapter appended a compact JSON Schema as a format contract to the base template; other model/run adapters differed and are retained in their archived configurations. These formatting instructions did not add clinical evidence or a gold answer. Raw output, strict-format validity, canonical validity after removal of an optional outer JSON fence, and validation errors were retained separately. We used five fresh invocations per exact input to assess response stability. Frozen run configurations record model revisions, rendered-prompt and input hashes, and the selected source passages.

For P6R, the selected 50 calls per model comprise 30 unchanged calls and 20 later calls using the source-reviewed independent passage for recognition and independent-update conditions. The superseded and selected inputs are not pooled. This compact provenance note preserves the reproducible selection without treating P6R as a distinct experimental design.

\subsection{Model configuration and compute}

The open-weight checkpoints were loaded from pinned Hugging Face revisions recorded in \texttt{configs/transformers\_gpu\_models.json}. The runs used Kaggle GPU sessions with two Tesla T4 GPUs. The archived configurations preserve exact software environments, model-specific loading choices, and runner settings. GPT-5.5 was accessed through Codex; its parameter count and serving hardware are not available in the archived records. Reliable wall-clock and GPU-hour totals were not retained.

The cross-model runs keep the clinical claims, passages, gold behaviors, and semantic task instructions fixed within each input version while adapting model-specific chat and formatting wrappers. The original ten-family suite and later four-family package were executed in separate stages.

\subsection{Call accounting}

Table~\ref{tab:call-accounting} reports the planned calls for each analysis block. Repeated invocations of an exact input measure response stability; they do not increase the number of clinical families. Secondary, diagnostic, and exploratory blocks are kept outside the 14-family primary/follow-up comparison.

\begin{table*}[t]
\centering
\small
\begin{tabular}{@{}lccc@{}}
\toprule
Block & Planned calls per tested model & Tested model conditions & Planned calls \\
\midrule
Original ten-family suite & 330 & 5 & 1,650 \\
Later four-family package & 220 & 5 & 1,100 \\
Prompt-order robustness & 40 & 3 & 120 \\
CF-P3 oracle + H3-M1 diagnostics & 65 & 3 & 195 \\
Exploratory lineage-by-overlap control & -- & model-dependent & 36 total \\
\bottomrule
\end{tabular}
\caption{Call accounting by analysis block. Completion does not imply validity; rejected outputs are reported separately in Appendix~\ref{app:validity}.}
\label{tab:call-accounting}
\end{table*}

\subsection{Artifact contents and rerun requirements}

The anonymized artifact referenced in the main paper contains the paper source, frozen inputs, rendered prompts, schemas, configurations,
source-audit records, raw and parsed outputs, manifests, content hashes, figures, and rerun scripts. GPT-5.5 calls were made through Codex with medium reasoning effort. The archived records support audit of these calls, while an exact rerun
requires Codex access. The open-weight conditions can be rerun from the pinned model revisions and recorded GPU runner.

\section{Original provenance results by family}
\label{app:provenance-detail}

Table~\ref{tab:original-provenance-detail} gives the recognition and same-trial aggregation rows behind the paper's central cross-task comparison. A rejected call receives no semantic score, so a $0/0$ cell is not a measured zero.


\begin{figure*}[p]
\centering
\begin{minipage}{0.96\textwidth}
\noindent\textbf{Comparative-recognition prompt (v2.2)}\par\smallskip

The same template handles provenance, applicability, and direction; \texttt{target\_relation} selects one question. Update A and Update B are also tested in the reverse order.

\begin{lstlisting}[style=paperprompt]
You are comparing two possible evidence updates for the same fixed medical claim and the same
initial evidence set.

Use only the claim, initial evidence, Update A, Update B, and displayed source metadata shown in
MODEL INPUT. Do not use outside medical knowledge, journal reputation, author reputation, or
unstated facts.

Answer only the relation named by target_relation. The matched comparison is between Update A and
Update B; it is not between the initial evidence and a single added item. Do not decide how the
final claim verdict should change.

If target_relation is provenance, answer this question:
Which update adds genuinely independent evidence relative to the initial evidence set?

For provenance, genuinely independent evidence means evidence derived from a distinct underlying
study, trial, or participant cohort. A different PMID, publication, follow-up report, subgroup
analysis, or secondary analysis of the same underlying study is not independent.

If target_relation is applicability, answer this question:
Which update is applicable to the claim as written?

If target_relation is direction, answer this question:
Which update introduces genuine opposing evidence relative to the claim?

Choose exactly one comparative_answer:
- A: Update A satisfies the target_relation question.
- B: Update B satisfies the target_relation question.
- no_meaningful_difference: Update A and Update B do not differ meaningfully on the target_relation
  question based on the supplied text and metadata.

Judge only the requested relation. For example, an update can report an agreeing direction while
being less applicable, or can be from the same trial while still reporting a relevant result.

For metadata_basis, list only displayed identifiers or source facts used for the comparison.

For every evidence_basis entry, copy one contiguous, character-for-character substring from the
corresponding passage_text. The evidence_quote must pass an exact substring check:
- preserve uppercase/lowercase exactly;
- preserve punctuation exactly;
- do not lowercase the first letter of a sentence;
- do not join separate spans;
- do not insert ellipses;
- if copying a phrase immediately after a heading such as METHODS: or RESULTS:, either include the
  heading exactly or copy the following sentence with its original capitalization.

An empty evidence_basis list is allowed when the supplied passages do not establish the comparison.

Return JSON only, matching the supplied response schema and its field order. Echo item_id and
target_relation exactly. Give one concise reason; do not provide hidden chain-of-thought. Do not
recommend or reveal a final verifier label or score.

MODEL INPUT:
{model_input_json}
\end{lstlisting}
\end{minipage}
\caption{Complete comparative-recognition prompt.}
\label{fig:recognition-prompt}
\end{figure*}

\begin{figure*}[p]
\centering
\begin{minipage}{0.96\textwidth}
\noindent\textbf{Document-level verifier prompt (v2)}\par\smallskip

The verifier scores the initial evidence when \texttt{added\_evidence} is null and the initial-plus-added state otherwise. It is not given the expected update relation.

\begin{lstlisting}[style=paperprompt]
You are assessing how well a fixed medical claim is grounded by the supplied evidence.

Use only the claim, evidence passages, and source metadata shown in MODEL INPUT. Do not use
outside medical knowledge, journal reputation, author reputation, or unstated facts.

If added_evidence is null, assess the initial evidence as the current evidence set. Otherwise,
assess the initial evidence and added evidence together as the current set. Assess every evidence
item in the displayed order. For each evidence_quote, copy one contiguous,
character-for-character substring from that evidence item's passage_text. Do not join separate
spans, change words, or insert ellipses.

Assess the claim as written. Do not weaken, strengthen, or reinterpret the claim.

Use these per-evidence labels:
- supports: the passage directly states or entails the complete claim as written.
- conflicts: the passage directly states or entails the opposite of a central part of the claim.
- insufficient: the passage concerns the claim but does not contain enough information to decide it.
- out_of_scope: the passage does not bear closely enough on the claim as written to determine
  whether it supports or conflicts with it.

Use these combined labels:
- supported: the supplied evidence grounds the complete claim.
- contradicted: the supplied evidence grounds an opposing claim.
- unsupported: the supplied evidence does not ground the claim.
- not_enough_evidence: the supplied evidence is relevant but insufficient for a decision.
- mixed_evidence: the supplied evidence contains substantive support and substantive conflict.

Score definitions:
- semantic_support_score: how directly an individual passage supports the complete claim, from
  0 (no support) to 100 (explicit complete support).
- warrant_strength_score: how strongly the supplied evidence justifies the complete claim as
  written, from 0 (no warrant) to 100 (complete warrant).
- combined_support_score: strength of the supporting side of the current evidence set, from 0 to 100.
- combined_conflict_score: strength of the opposing side of the current evidence set, from 0 to 100.
- confidence: confidence in this assessment, not confidence that the medical claim is true.

Return JSON only, matching the supplied response schema and its field order. Echo item_id exactly.
Give concise reasons; do not provide hidden chain-of-thought.

MODEL INPUT:
{model_input_json}
\end{lstlisting}
\end{minipage}
\caption{Complete neutral document-level verifier prompt.}
\label{fig:document-prompt}
\end{figure*}

\begin{figure*}[p]
\centering
\begin{minipage}{0.96\textwidth}
\noindent\textbf{Study-grouped verifier prompt (v1)}\par\smallskip

This is a fresh verifier call over the same displayed evidence; the third paragraph adds the study-level aggregation instruction.

\begin{lstlisting}[style=paperprompt]
You are assessing how well a fixed medical claim is grounded by the supplied evidence.

Use only the claim, evidence passages, and source metadata shown in MODEL INPUT. Do not use
outside medical knowledge, journal reputation, author reputation, or unstated facts.

This is the study-level provenance-aware aggregation condition. Before making the combined
judgment, group evidence items by their underlying study identity using trial_id and source_group_id.
Multiple reports from the same study may add detail about that study, but they must not be counted as
independent corroborating studies. Evidence from a different trial_id may count as independent
corroboration if it is applicable to the claim and points in the relevant direction.

If added_evidence is null, assess the initial evidence as the current evidence set. Otherwise,
assess the initial evidence and added evidence together as the current set. Assess every evidence
item in the displayed order. For each evidence_quote, copy one contiguous,
character-for-character substring from that evidence item's passage_text. Do not join separate
spans, change words, or insert ellipses.

Assess the claim as written. Do not weaken, strengthen, or reinterpret the claim.

Use these per-evidence labels:
- supports: the passage directly states or entails the complete claim as written.
- conflicts: the passage directly states or entails the opposite of a central part of the claim.
- insufficient: the passage concerns the claim but does not contain enough information to decide it.
- out_of_scope: the passage does not bear closely enough on the claim as written to determine
  whether it supports or conflicts with it.

Use these combined labels:
- supported: the supplied evidence grounds the complete claim.
- contradicted: the supplied evidence grounds an opposing claim.
- unsupported: the supplied evidence does not ground the claim.
- not_enough_evidence: the supplied evidence is relevant but insufficient for a decision.
- mixed_evidence: the supplied evidence contains substantive support and substantive conflict.

Score definitions:
- semantic_support_score: how directly an individual passage supports the complete claim, from
  0 (no support) to 100 (explicit complete support).
- warrant_strength_score: how strongly the supplied evidence justifies the complete claim as
  written, from 0 (no warrant) to 100 (complete warrant).
- combined_support_score: strength of the supporting side of the current evidence set, from 0 to 100.
- combined_conflict_score: strength of the opposing side of the current evidence set, from 0 to 100.
- confidence: confidence in this assessment, not confidence that the medical claim is true.

Return JSON only, matching the supplied response schema and its field order. Echo item_id exactly.
Give concise reasons; do not provide hidden chain-of-thought.

MODEL INPUT:
{model_input_json}
\end{lstlisting}
\end{minipage}
\caption{Complete study-grouped verifier prompt.}
\label{fig:grouped-prompt}
\end{figure*}




\begin{table*}[t]
\centering
\small
\begin{tabular}{@{}llcc@{}}
\toprule
Model & Family & Recognition correct/valid & Same-trial over-credit/valid \\
\midrule
GPT-5.5 & CF-P1 & 10/10 & 5/5 \\
          & CF-P2 & 10/10 & 5/5 \\
          & CF-P3 & 10/10 & 5/5 \\
\addlinespace[2pt]
Qwen3-4B & CF-P1 & 10/10 & 5/5 \\
         & CF-P2 & 10/10 & 0/0 (5 rejected) \\
         & CF-P3 & 10/10 & 5/5 \\
\addlinespace[2pt]
Gemma3-4B & CF-P1 & 5/10 & 5/5 \\
          & CF-P2 & 5/10 & 0/0 (5 rejected) \\
          & CF-P3 & 5/10 & 0/0 (5 rejected) \\
\addlinespace[2pt]
Llama-3.1-8B & CF-P1 & 5/10 & 0/0 (5 rejected) \\
              & CF-P2 & 5/10 & 0/0 (5 rejected) \\
              & CF-P3 & 5/10 & 5/5 \\
\addlinespace[2pt]
Ministral-3-8B & CF-P1 & 10/10 & 5/5 \\
                 & CF-P2 & 5/5 (5 rejected) & 0/0 (5 rejected) \\
                 & CF-P3 & 5/5 (5 rejected) & 5/5 \\
\bottomrule
\end{tabular}
\caption{Recognition and same-trial aggregation for every model in CF-P1--P3. Rejected calls are shown explicitly; 0/0 has no valid semantic estimate.}
\label{tab:original-provenance-detail}
\end{table*}

\section{Study-grouping and later-family results}
\label{app:later-detail}

Table~\ref{tab:later-detail} keeps P5R/P6R recognition, document-level over-credit, grouped over-credit, and the independent-trial preservation check together. In panel A, an \texttt{independent\_corroboration} flag on the same-trial update is attributable over-credit. Panel B asks whether the intervention still preserves support for a genuinely independent trial. The H5/H6 results are reported separately because both compositional baselines already contain an independent study; their global corroboration flag cannot isolate credit assigned specifically to the newly added same-trial report.

\begin{table*}[t]
\centering
\small
\textit{A. Recognition and same-trial over-credit}\par\smallskip
\begin{tabular}{@{}llccc@{}}
\toprule
Family & Model & Recognition correct/valid (rej.) & D over-credit/valid (rej.) & G over-credit/valid (rej.) \\
\midrule
P5R & GPT-5.5 & 10/10 (0) & 4/5 (0) & 0/5 (0) \\
    & Qwen3-4B & 10/10 (0) & 5/5 (0) & 5/5 (0) \\
    & Gemma3-4B & 5/10 (0) & 5/5 (0) & 5/5 (0) \\
    & Llama-3.1-8B & 5/10 (0) & 0/5 (0) & 0/5 (0) \\
    & Ministral-3-8B & 10/10 (0) & 0/0 (5) & 0/0 (5) \\
\addlinespace[2pt]
P6R & GPT-5.5 & 10/10 (0) & 4/5 (0) & 0/5 (0) \\
    & Qwen3-4B & 10/10 (0) & 5/5 (0) & 5/5 (0) \\
    & Gemma3-4B & 5/5 (5) & 5/5 (0) & 5/5 (0) \\
    & Llama-3.1-8B & 5/10 (0) & 5/5 (0) & 0/0 (5) \\
    & Ministral-3-8B & 10/10 (0) & 5/5 (0) & 0/5 (0) \\
\bottomrule
\end{tabular}
\par\medskip
\textit{B. Independent-trial support: document-level $\to$ study-grouped}\par\smallskip
\noindent\emph{Panel B is intentionally compact: all valid rows are unchanged at $5\to5$ except Ministral on P5R, whose grouped condition has no valid output.}\par\smallskip
\begin{tabular}{@{}lcc@{}}
\toprule
Model & P5R & P6R \\
\midrule
GPT-5.5 & $5/5\to5/5$ & $5/5\to5/5$ \\
Qwen3-4B & $5/5\to5/5$ & $5/5\to5/5$ \\
Gemma3-4B & $5/5\to5/5$ & $5/5\to5/5$ \\
Llama-3.1-8B & $5/5\to5/5$ & $5/5\to5/5$ \\
Ministral-3-8B & $5/5\to0/0$ & $5/5\to5/5$ \\
\bottomrule
\end{tabular}
\caption{Pure-provenance P5R/P6R results. Recognition and aggregation are separate stateless calls. Panel A counts same-trial over-credit among valid outputs; panel B counts support for the independent-trial update (flags/valid calls). In either panel, $0/0$ means five rejected outputs, not a measured zero or loss of support.}
\label{tab:later-detail}
\end{table*}

\begin{table*}[t]
\centering
\small
\begin{tabular}{@{}llccc@{}}
\toprule
Family & Model & Recognition correct/valid (rej.) & D global flag/valid (rej.) & G global flag/valid (rej.) \\
\midrule
H5 & GPT-5.5 & 20/20 (0) & 5/5 (0) & 4/5 (0) \\
   & Qwen3-4B & 15/20 (0) & 5/5 (0) & 5/5 (0) \\
   & Gemma3-4B & 10/20 (0) & 0/0 (5) & 0/0 (5) \\
   & Llama-3.1-8B & 10/20 (0) & 5/5 (0) & 0/0 (5) \\
   & Ministral-3-8B & 20/20 (0) & 0/0 (5) & 0/0 (5) \\
\addlinespace[2pt]
H6 & GPT-5.5 & 10/10 (0) & 5/5 (0) & 1/5 (0) \\
   & Qwen3-4B & 10/10 (0) & 5/5 (0) & 5/5 (0) \\
   & Gemma3-4B & 5/10 (0) & 5/5 (0) & 5/5 (0) \\
   & Llama-3.1-8B & 5/10 (0) & 5/5 (0) & 0/0 (5) \\
   & Ministral-3-8B & 10/10 (0) & 5/5 (0) & 5/5 (0) \\
\bottomrule
\end{tabular}
\caption{Compositional H5/H6 results. D and G count the global \texttt{independent\_corroboration} flag on a same-trial update, not attributable over-credit: both baselines contain an independent study. H5 recognition includes provenance and applicability comparisons. A $0/0$ cell has no valid semantic estimate.}
\label{tab:compositional-detail}
\end{table*}

\section{Applicability, direction, and compositional controls}
\label{app:control-axis-results}

These analyses are specificity controls. Table~\ref{tab:control-axis-recognition} gives comparative recognition for the six non-provenance primitive families and H3. In CF-A1--A3 the comparison concerns clinical applicability; in CF-D1--D3 it concerns result direction. H3 combines provenance, applicability, and direction, so its total is not a pure lineage score.

\begin{table*}[t]
\centering
\small
\setlength{\tabcolsep}{3.5pt}
\begin{tabular}{@{}lccccccc@{}}
\toprule
Model & CF-A1 & CF-A2 & CF-A3 & CF-D1 & CF-D2 & CF-D3 & H3 \\
\midrule
GPT-5.5        & 10/10 & 10/10 & 10/10 & 10/10 & 10/10 & 10/10 & 30/30 \\
Qwen3-4B       & 10/10 & 10/10 & 10/10 & $0/5^{\dagger}$ & 5/10 & 10/10 & 30/30 \\
Gemma3-4B      & 5/10 & 5/10 & 10/10 & 5/10 & $5/5^{\dagger}$ & 5/10 & 15/30 \\
Llama-3.1-8B   & 5/10 & $0/5^{\dagger}$ & 5/10 & 5/10 & 5/10 & 5/10 & 15/30 \\
Ministral-3-8B & 10/10 & 10/10 & 10/10 & 10/10 & 5/10 & 10/10 & 20/30 \\
\bottomrule
\end{tabular}
\caption{Comparative recognition in applicability (A), direction (D), and compositional (H3) families. Cells are correct/valid calls. Each primitive family has ten planned recognition calls; H3 has 30 across three relations. $\dagger$ marks five additional rejected recognition calls, not five incorrect answers. Repetitions are within-input stability checks.}
\label{tab:control-axis-recognition}
\end{table*}

The semantic integration outcomes are shown separately below. Unless noted otherwise, the family-level applicability and direction integration rows are GPT-5.5 medium results; the open-weight primary package does not provide a harmonized family-by-family semantic integration summary for these two control axes, so no cross-model integration rate is inferred from them.

\begin{table*}[t]
\centering
\small
\begin{tabularx}{\textwidth}{@{}l l X X@{}}
\toprule
Control & Families/conditions & Expected handling & GPT-5.5 result \\
\midrule
Exact repeat / null & CF-P1--P3, CF-A1--A3, CF-D1--D3 & no new independent evidence & duplicate or unchanged in 45/45 calls \\
Same-trial update & CF-P1--P3 & no independent corroboration & over-credit in 15/15 calls \\
Independent update & P5R, P6R & preserve independent support & 5/5 $\to$ 5/5 in each family \\
Out-of-scope positive & CF-A1--A3 & ignore for the fixed claim & out\_of\_scope and ignored, 15/15 \\
Challenging update & CF-D1--D3 & retain support and register conflict & conflict or mixed evidence, 15/15 \\
\bottomrule
\end{tabularx}
\caption{Compact control-outcome summary for the approved confirmatory and later pure-provenance conditions. The 45 exact-repeat calls are nine primitive families with five repetitions each; the 15/15 same-trial, out-of-scope, and challenging-update entries are three families with five repetitions each. The independent-support row is the separate P5R/P6R safety check.}
\label{tab:control-outcomes}
\end{table*}

\begin{table*}[t]
\centering
\small
\begin{tabularx}{\textwidth}{@{}l l X@{}}
\toprule
Family & Target update & GPT-5.5 integration outcome \\
\midrule
CF-A1 & applicable vs positive out of scope & applicable update used; out-of-scope update ignored, 5/5 \\
CF-A2 & applicable vs positive out of scope & applicable update used; out-of-scope update ignored, 5/5 \\
CF-A3 & first-line vs relapsed maintenance & applicable update used; out-of-scope update marked out\_of\_scope and ignored, 5/5 \\
CF-D1 & agreeing vs mortality-challenging ICU evidence & challenging update treated as conflict/mixed, 5/5 \\
CF-D2 & agreeing vs mortality-challenging diabetes evidence & challenging update treated as conflict/mixed, 5/5 \\
CF-D3 & agreeing vs annual high-dose vitamin-D evidence & agreeing update supports 5/5; challenging update conflicts and yields mixed\_evidence, 5/5 \\
\bottomrule
\end{tabularx}
\caption{Family-level applicability and direction integration for GPT-5.5. These are separate neutral-verifier calls from the comparative-recognition calls; the table is a control-axis audit, not evidence that applicability or direction is universally solved.}
\label{tab:integration-controls}
\end{table*}

In the held-out H3 family, GPT-5.5 and Qwen each answer all 30 comparative recognition calls correctly across provenance, applicability, and direction. In separate aggregation calls, however, the added same-trial WOSCOPS report receives an independent-corroboration flag in 4/5 and 5/5 valid calls, respectively. The 30 recognition calls comprise ten comparisons for each of the three relations.

\section{Secondary robustness and exploratory lineage-by-overlap control}
\label{app:secondary-analyses}

\subsection{Prompt-order robustness}

The secondary prompt audit reran the P5R/P6R document-level conditions with a semantically equivalent reordered verifier prompt. It contains 40 calls per model for GPT-5.5, Qwen3-4B, and Gemma3-4B: two families, same-trial and independent updates, two prompt variants, and five repetitions. GPT-5.5 marks same-trial updates as independent in 16/20 records and passes the independent-update semantic audit in 14/20. Qwen marks every same-trial record as independent (20/20) and passes 15/20 independent-update audits. Gemma is canonically parseable in all 40 records but strict-valid in 0/40 because each response has one outer Markdown JSON fence; its same-trial records fail the provenance flag rule in 20/20, while the independent-update audit passes 20/20. These calls test robustness to the two recorded prompt variants only and are not pooled with the primary denominators.

\begin{table*}[t]
\centering
\small
\begin{tabular}{@{}lcccc@{}}
\toprule
Model & Same-trial over-credit & Independent audit pass & Strict valid & Canonical valid \\
\midrule
GPT-5.5 & 16/20 & 14/20 & 40/40 & 40/40 \\
Qwen3-4B & 20/20 & 15/20 & 40/40 & 40/40 \\
Gemma3-4B & 20/20 & 20/20 & 0/40 & 40/40 \\
\bottomrule
\end{tabular}
\caption{Secondary prompt-order robustness audit on P5R/P6R. Each model has 40 calls across two prompt variants, two families, same-trial and independent updates, and five repetitions. Same-trial over-credit is a forbidden \texttt{independent\_corroboration} flag; the independent audit requires both the flag and use of the added independent evidence. Gemma's canonical validity reflects removal of one outer Markdown JSON fence; raw outputs remain unchanged.}
\label{tab:prompt-robustness}
\end{table*}

\subsection{Lineage-by-overlap construction and results}

The exploratory control uses one source-reviewed ProMISe/ARISE family. The fixed claim concerns 90-day all-cause mortality for early septic shock under early goal-directed therapy versus usual resuscitation. The initial anchor is the ProMISe primary report (\texttt{ISRCTN36307479}); the same-trial update is the ProMISe HTA report with the same registry identifier; the independent update is the ARISE primary report (\texttt{NCT00975793}). The selected passages are source-faithful result or conclusion passages. Audit records preserve registry linkage, clinical scope, source locator, exact text, and passage hashes. No passage was synthetically paraphrased, trimmed, or padded, and the model-visible input does not contain \texttt{source\_group\_id}.

We measured overlap against the initial ProMISe numerical-result anchor using lowercase Unicode-normalized word/number tokens with punctuation removed. We computed token-set Jaccard, ROUGE-L F1, and cosine similarity from the pinned \texttt{sentence-transformers/all-mpnet-base-v2} revision \texttt{e8c3b32e\allowbreak df5434bc\allowbreak 2275fc9b\allowbreak ab85f826\allowbreak 40a19130}. Within each lineage, the high-overlap passage exceeded the low-overlap passage on all three measures, with a cosine gap of at least 0.10 and a lexical gap of at least 0.10 on Jaccard or ROUGE-L. We also checked the matched high and low cells for cross-lineage comparability. These thresholds define overlap strata; estimating a causal overlap effect requires a separately designed intervention.

\begin{table*}[t]
\centering
\small
\setlength{\tabcolsep}{3pt}
\begin{tabularx}{\textwidth}{@{}l >{\raggedright\arraybackslash}X r r c c c@{}}
\toprule
Cell & Source passage & Tokens & Chars & Jaccard & ROUGE-L & Cosine \\
\midrule
Anchor & ProMISe primary numerical result & 68 & 335 & -- & -- & -- \\
Same/high & ProMISe HTA numerical result & 57 & 299 & 0.634615 & 0.400000 & 0.846106 \\
Same/low & ProMISe HTA abstract conclusion & 30 & 182 & 0.109375 & 0.122449 & 0.656978 \\
Independent/high & ARISE primary numerical result & 56 & 303 & 0.423729 & 0.370968 & 0.922640 \\
Independent/low & ARISE primary conclusion & 23 & 132 & 0.118644 & 0.109890 & 0.678653 \\
\bottomrule
\end{tabularx}
\caption{Frozen overlap measurements for the one-family policy-instructed lineage control. Scores are computed against the initial ProMISe numerical-result anchor; ``high'' and ``low'' are within-lineage strata, not causal treatments.}
\label{tab:overlap-construction}
\end{table*}

The verifier prompt for this control explicitly states that another report from the same trial is non-independent and a different trial may provide independent corroboration. The control evaluates policy application across overlap strata; causal overlap effects and unprompted lineage discovery require separate designs. GPT-5.5 produces the expected combination in all four cells (20/20 calls); other model conditions are mixed, as described in the archived analysis. Table~\ref{tab:overlap-policy-results} reports the policy-defined GPT-5.5 outcomes.

\begin{table*}[t]
\centering
\small
\begin{tabular}{@{}lccc@{}}
\toprule
Cell & Expected relation & Duplicate evidence & Independent corroboration \\
\midrule
Same/high & same trial & 5/5 & 0/5 \\
Same/low & same trial & 5/5 & 0/5 \\
Independent/high & independent trial & 0/5 & 5/5 \\
Independent/low & independent trial & 0/5 & 5/5 \\
\bottomrule
\end{tabular}
\caption{Policy-defined lineage-by-overlap results for GPT-5.5. All 20 calls labeled the claim supported. The expected pattern is duplicate evidence for same-trial passages and independent corroboration for independent-trial passages. The four cells each contain five fresh stateless calls. This one-family result reports policy application across overlap strata. Separate designs are needed to estimate overlap effects or unprompted lineage discovery.}
\label{tab:overlap-policy-results}
\end{table*}

\section{Post-hoc oracle and cue-ablation diagnostics}
\label{app:diagnostic-results}

These diagnostics followed the primary provenance failure and were frozen before their model calls. In CF-P3, the relation oracle supplies the item-specific study relation, whereas the policy oracle states that repeated or same-trial reports can add detail but not independent corroboration. Each has exact-repeat, same-trial, and independent-study conditions (30 responses per tested model).

H3-M1 removes only \texttt{source\_group\_id} from the cue-rich H3 setup while preserving the claim, passages, evidence order, titles, trial identifiers, and other metadata. It contributes two counterbalanced provenance-recognition inputs and five neutral-update inputs, or 35 responses per tested model. Together, the CF-P3 oracle block and H3-M1 contain 65 responses per tested model. GPT-5.5, Qwen3-4B, and Gemma3-4B completed this diagnostic design; Llama and Ministral were not run in this block. Completion is distinct from output validity.

In CF-P3, both GPT-5.5 oracles remove the same-trial flag in 5/5 valid calls and preserve the independent-trial control. Qwen retains the flag under both oracles, and Gemma's same-trial oracle outputs are rejected. In H3-M1, GPT-5.5 retains 10/10 provenance recognition but flags the same-trial report in 5/5 verifier calls. Full denominators follow.

\begin{table*}[t]
\centering
\small
\setlength{\tabcolsep}{4pt}
\begin{tabular}{@{}lccccc@{}}
\toprule
 & \multicolumn{3}{c}{CF-P3 oracles} & \multicolumn{2}{c}{H3-M1 cue ablation} \\
\cmidrule(lr){2-4}\cmidrule(l){5-6}
Model & Relation: same & Policy: same & Independent R/P & Recognition & Same-trial flag \\
\midrule
GPT-5.5 & 0/5 & 0/5 & 5/5, 5/5 & 10/10 & 5/5 \\
Qwen3-4B & 5/5 & 5/5 & 5/5, 5/5 & 5/10 & 5/5 \\
Gemma3-4B & 0/0 (5 rej.) & 0/0 (5 rej.) & 5/5, 5/5 & 5/10 & 5/5 \\
\bottomrule
\end{tabular}
\caption{Post hoc diagnostic outcomes, not additional primary families. R/P denotes the independent-trial condition under the relation and policy oracles, respectively. The two CF-P3 ``same'' columns and the H3-M1 same-trial column count the independent-corroboration flag on a same-trial update. A $0/0$ cell has five rejected outputs and no semantic estimate. Qwen has five further rejected H3-M1 exact-repeat outputs; Gemma has 25 rejected diagnostic outputs overall. GPT-5.5 has no archived error row in this block.}
\label{tab:diagnostic-results}
\end{table*}

\section{Validity and rejection accounting}
\label{app:validity}

Table~\ref{tab:family-validity} reports canonical validity by model and family. Cells show valid/rejected calls, not correctness; rejected responses receive no semantic score. Repeated calls assess within-input stability, while the 14 clinical families are the empirical units.

\begin{table*}[!b]
\centering
\footnotesize
\begin{tabular}{@{}lrrrrrr@{}}
\toprule
Family & Planned & GPT-5.5 & Qwen3-4B & Gemma3-4B & Llama-3.1-8B & Ministral-3-8B \\
\midrule
CF-P1 & 30 & 30/0 & 30/0 & 15/15 & 20/10 & 30/0 \\
CF-P2 & 30 & 30/0 & 10/20 & 10/20 & 20/10 & 10/20 \\
CF-P3 & 30 & 30/0 & 25/5 & 10/20 & 25/5 & 25/5 \\
CF-A1 & 30 & 30/0 & 30/0 & 25/5 & 25/5 & 30/0 \\
CF-A2 & 30 & 30/0 & 30/0 & 20/10 & 25/5 & 30/0 \\
CF-A3 & 30 & 30/0 & 20/10 & 15/15 & 10/20 & 20/10 \\
CF-D1 & 30 & 30/0 & 15/15 & 20/10 & 10/20 & 30/0 \\
CF-D2 & 30 & 30/0 & 30/0 & 15/15 & 25/5 & 10/20 \\
CF-D3 & 30 & 30/0 & 25/5 & 25/5 & 10/20 & 25/5 \\
H3    & 60 & 60/0 & 55/5 & 55/5 & 40/20 & 55/5 \\
\midrule
P5R & 50 & 50/0 & 45/5 & 50/0 & 50/0 & 25/25 \\
P6R & 50 & 50/0 & 50/0 & 45/5 & 40/10 & 45/5 \\
H5  & 70 & 70/0 & 65/5 & 45/25 & 60/10 & 30/40 \\
H6  & 50 & 50/0 & 50/0 & 50/0 & 35/15 & 45/5 \\
\midrule
Total & 550 & 550/0 & 480/70 & 400/150 & 395/155 & 410/140 \\
\bottomrule
\end{tabular}
\caption{Canonical-valid/rejected calls in the selected 14-family comparison. Cells record response-contract validity rather than correctness; rejected calls receive no semantic score. Planned calls are \emph{per model}; each cell sums to its row's planned count.}
\label{tab:family-validity}
\end{table*}

\end{document}